\documentclass[conference]{IEEEtran}
\IEEEoverridecommandlockouts
\usepackage{cite}
\usepackage{amsmath,amssymb,amsfonts}
\usepackage{algorithmic}
\usepackage{graphicx}
\usepackage{textcomp}
\usepackage{xcolor}

\usepackage{pgf}
\usepackage{bm}
\usepackage{mathtools}
\usepackage[linesnumbered, ruled, vlined]{algorithm2e}
\usepackage[shortlabels]{enumitem}
\usepackage{booktabs}
\usepackage[makeroom]{cancel}
\usepackage{cite}
\usepackage{hyperref}

\DeclareMathOperator*{\argmin}{arg\,min}

\def\BibTeX{{\rm B\kern-.05em{\sc i\kern-.025em b}\kern-.08em
    T\kern-.1667em\lower.7ex\hbox{E}\kern-.125emX}}
\begin{document}
\bstctlcite{IEEEexample:BSTcontrol}

\title{Learning Individualized Treatment Rules with Estimated Translated Inverse Propensity Score\\
}

\author{
\IEEEauthorblockN{
Zhiliang Wu\textsuperscript{1, 2}, 
Yinchong Yang\textsuperscript{1}, 
Yunpu Ma\textsuperscript{1, 2}, 
Yushan Liu\textsuperscript{1, 2}, 
Rui Zhao\textsuperscript{1, 2},  
Michael Moor\textsuperscript{3}, 
Volker Tresp\textsuperscript{1, 2}
}
\IEEEauthorblockA{
\textsuperscript{1}Siemens AG, 
\textsuperscript{2}LMU Munich, 
Munich, Germany,
\{firstname.lastname\}@siemens.com}
\IEEEauthorblockA{
\textsuperscript{3}ETH Zurich, Basel, Switzerland, 
michael.moor@bsse.ethz.ch
}
}

\maketitle

\begin{abstract}
Randomized controlled trials typically analyze the effectiveness of treatments with the goal of making treatment recommendations for patient subgroups. 
With the advance of electronic health records, a great variety of data has been collected in clinical practice, enabling the evaluation of treatments and treatment policies based on observational data.
In this paper, we focus on learning individualized treatment rules (ITRs) to derive a treatment policy that is expected to generate a better outcome for an individual patient. 
In our framework, we cast ITRs learning as a contextual bandit problem and minimize the expected risk of the treatment policy.
We conduct experiments with the proposed framework both in a simulation study and based on a real-world dataset. In the latter case, we apply our proposed method to learn the optimal ITRs for the administration of intravenous (IV) fluids and vasopressors (VP). 
Based on various offline evaluation methods, we could show that the policy derived in our framework demonstrates better performance compared to both the physicians and other baselines, including a simple treatment prediction approach. 
As a long-term goal, our derived policy might eventually lead to better clinical guidelines for the administration of IV and VP.
\end{abstract}


\section{Introduction}
Since the introduction of electronic health records (EHRs), machine learning has increasingly been used to analyze observational clinical data with the goal of individualizing patient care \cite{tresp2016going}. 
Compared to traditional rule-based strategies, where all patients with a specific disease in a particular patient group receive similar treatments, the goal of modern personalized medicine is to offer better care to individual patients, taking into account their heterogeneous characteristics. Personalized medicine might be especially important for situations where high-dimensional longitudinal data needs to be analyzed under time pressure, as in an emergency room (ER) or an intensive care unit (ICU). Here, treatment decisions might have to be made without the best medical expert for the case being readily available.

In personalized medicine, individualized treatment rules (ITRs) assign a treatment from a range of possible treatments to an individual patient based on his or her clinical characteristics \cite{zhou2017residual}. 
Ideally, all patients would have positive outcomes after receiving the treatments suggested by the optimal ITRs. In practice, one is interested in the ITRs' best mean performance. However, the evaluation of ITRs remains challenging, as it is unethical or even dangerous to apply newly learned rules directly to patients. Offline evaluation is the most widely used approach for such tasks. 
When learning the optimal ITRs, it is implicitly assumed that individualization can lead to better outcomes compared to current guidelines. In clinical practice, physicians might already perform some form of individualization by taking into account patient attributes that are not considered in the guidelines. In \textit{predictive modeling}, one attempts to directly copy the physicians' decision processes by using machine learning \cite{choi2016doctor}, which serves as one of our baseline methods. 

\begin{figure}[!htb]
    \centering
    \includegraphics[width=0.48\textwidth]{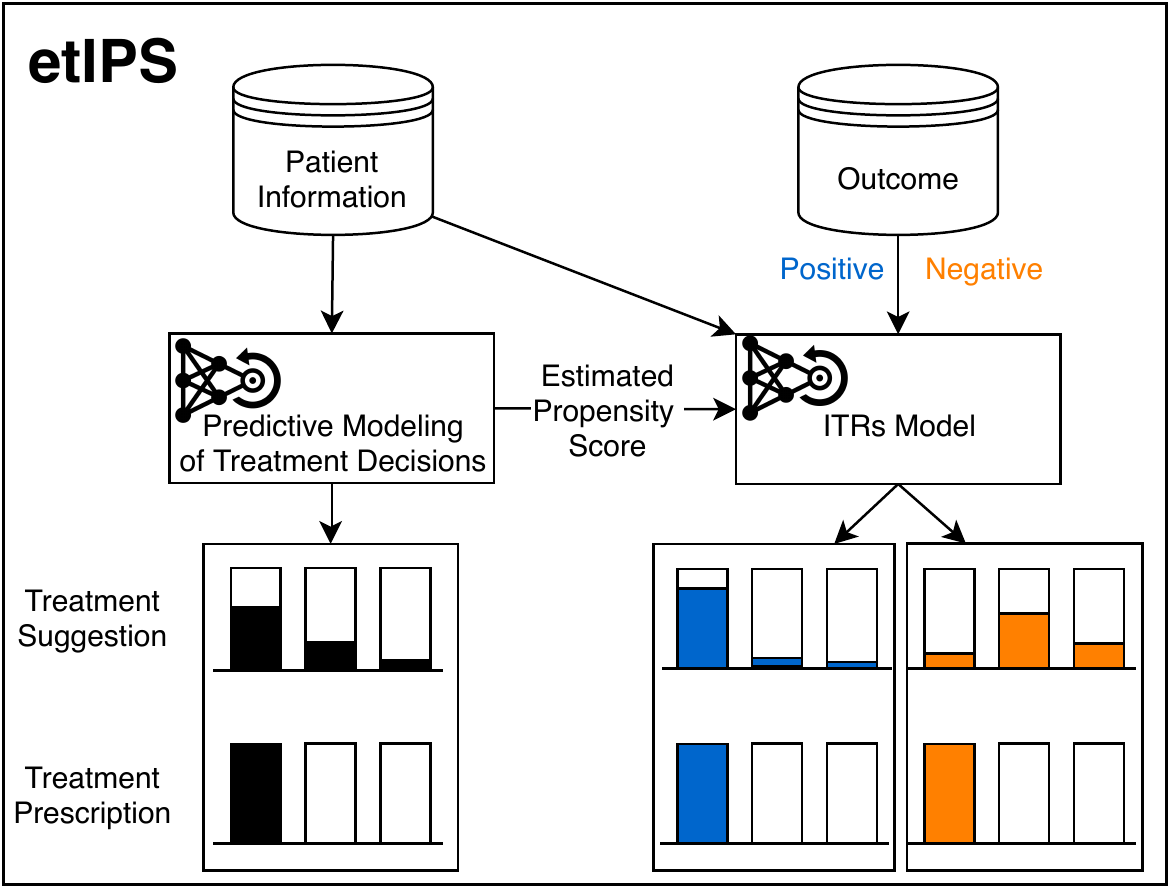}
    \caption{\textbf{etIPS} for learning the optimal ITRs: Both the predictive model (left) and the ITRs model (right) generate treatment suggestions based on available patient information. The predictive model is trained to mimic the physicians' decisions as well as possible. If the predictive model is trained to output probabilistic scores, it essentially estimates propensity scores. The ITRs are trained by encouraging treatments with a positive outcome as well as discouraging treatments with a negative outcome.  }
    \label{fig: framework}
\end{figure}

Recently, many researchers have built powerful machine learning models to predict the physicians' treatment decisions with neural networks\cite{esteban2015predicting, esteban2016predicting, yang2017predictive}. In particular, recurrent neural networks (and their advanced variants) are the de facto choice when dealing with sequential EHRs. In this paper, we show that recurrent neural networks are also suitable for learning the optimal ITRs within the proposed framework shown in Fig.~\ref{fig: framework}.  

From a machine learning perspective, the task of learning optimal ITRs can be formulated as treatment policy optimization based on the observed treatments and their received outcomes for individual patients. Such formulation is closely related to the contextual bandit problem, which concerns decision making in an environment where feedback is received only for a chosen action under a specific context. The challenge lies in the fact that only the feedback of an assigned action is observed, while the feedback of other actions remains unknown. Most work on the contextual bandit problem in machine learning concerns online services like content recommendations, where the context is a user's profile of interests in different topics, the action is the recommended item, and the feedback is the click action for the recommended item\cite{dudik2011doubly}. Online systems also record the model's assigned probability for each recommended item, which plays an essential role in learning a better policy. In the setup of clinical trials, the context can be viewed as the health level and treatment history of patients, the action refers to the treatment decision, and the feedback is the outcome observed after that specific treatment. The probability of assigning a particular treatment to the patient based on his or her covariates is known as the \textit{propensity score}\cite{rosenbaum1983central}. In randomized clinical trials, the propensity score is usually predefined for the experiments (e.g., 50\% for binary randomized clinical trials). However, in observational studies, the propensity score can only be estimated since it is implicit in the observed medical decisions. Many previous studies focus on the learning of ITRs with the predefined propensity scores\cite{qian2011performance, zhao2012estimating, zhou2017residual}.

Our contribution in this manuscript is threefold:
\begin{enumerate}
\item Inspired by previous works in predictive modeling of treatment decisions and contextual bandit problems, we present a general framework, etIPS, for learning ITRs based on sequential EHRs from observational studies by estimating the underlying true propensity score. 
\item With experiments on two simulated sequential classification tasks, we empirically verify that the estimated propensity score can replace the true propensity score for learning a better policy in contextual bandit problems.
\item We apply the proposed framework to the MIMIC-III dataset\cite{johnson2016mimic} to learn the optimal ITRs for the administration of intravenous (IV) fluids and vasopressors (VP). In various offline evaluations, the ITRs derived from our proposed method show better performance when compared to the physicians' decisions and other baselines. 
\end{enumerate}

\section{Related work}\label{sec: related}

\textbf{Predictive modeling with sequential EHRs}:
Recurrent neural networks (RNNs) have achieved great success on tasks such as machine translation in natural language processing\cite{bengio2003neural, mikolov2010recurrent, kim2016character,zhao2018learning,zhao2018efficient}. 
In machine translation, sentences are composed of variable number of words, just as EHRs consist of medical events of variable length. Esteban et al. have applied the sequence-to-sequence structure\cite{sutskever2014sequence} to predict clinical events of patients suffering from kidney failure \cite{esteban2016predicting}. In their work, the static EHRs are integrated into the network to achieve better performance. Meanwhile, Choi et al. propose \textit{Doctor AI} to predict diagnosis and medication prescriptions simultaneously\cite{choi2016doctor}. Furthermore, Choi et al. augment the network with attention mechanisms to improve both the accuracy and model interpretability\cite{choi2016retain}. More recently, Yang et al. have proposed to apply the many-to-one structure to predict the therapy decision for breast cancer\cite{yang2017predictive}, which only outputs one prediction for a sequence of events. However, predictive modeling is solely trained to mimic treatment decisions without taking into account the outcome information. 

\textbf{Learning Individualized Treatment Rules (ITRs)}: 
The learning of ITRs has attracted much attention in medical research. To get the best average outcome, Qian et al. propose a two-step method\cite{qian2011performance}. First, an outcome prediction model is fitted with the patient information and treatments. Second, ITRs are derived by selecting the treatment that promises to lead to the best outcome according to the trained model.  This approach relies heavily on the correctness of the outcome prediction model. In comparison, Zhao et al. propose the framework of outcome weighted learning (OWL) to construct a model that directly optimizes the outcome without learning an explicit outcome model\cite{zhao2012estimating}. In OWL, the learning of ITRs is formulated as a weighted classification problem and is solved by support vector machines. More recently, Zhou et al. have proposed the residual weighted learning (RWL) to improve the robustness of the ITRs learned by OWL\cite{zhou2017residual}. A separate regression model is fitted to estimate the baseline to compute the residual from the outcome. The discussed frameworks mainly focus on linear models and linear classifiers.  

\textbf{Learning the administration of IV and VP}: Komorowski et al. propose a reinforcement learning agent to learn the optimal strategies for sepsis management\cite{komorowski2018artificial}. A k-means algorithm is used to infer the states of the patients, 25 actions are defined by discretizing the dosage of IV and VP, and the mortality is used to define the long-term reward. The optimal policy is derived by solving a Markov decision process with policy iteration. However, mortality is a sparse and noisy long-term reward for both learning and evaluation. In this paper, we have a similar problem setting, but take advantage of an immediate reward to learn and evaluate the optimal ITRs. 

\textbf{Batch learning from bandit feedback (BLBF)}:
Bandit learning is commonly applied in online recommendation systems, where algorithms are evolving by trial and error with real-time feedback from users. In medical applications,  it is more common to train algorithms offline, mostly for safety considerations. 
Batch learning from bandit feedback is one of the offline versions of the contextual bandit problem, where the algorithm is trained with a batch of bandit feedback without online interactions\cite{beygelzimer2009offset}. 
Under the BLBF setting, the two-step method of deriving an optimal decision by maximizing the best estimated outcome proposed by Qian et al. is called the \textit{Direct Method} (DM), whereas the approaches to optimize weighted outcomes directly proposed by Zhao et al. are known as \textit{Inverse Propensity Score} (IPS) methods \cite{dudik2011doubly}. Swaminathan et al. cast BLBF as a counterfactual risk minimization problem. They propose the Policy Optimizer for Exponential Models (POEM) to improve the robustness of IPS methods\cite{swaminathan2015counterfactual}. Besides, Swaminathan et al. propose to use the self-normalized estimator for counterfactual learning (Norm-POEM) to alleviate the propensity score overfitting problem \cite{swaminathan2015self}. Both POEM and Norm-POEM are only applicable to linear models. More recently, Joachim et al. have proposed to reformulate the self-normalized estimator to train neural networks with bandit feedback\cite{joachims2018deep}. However, all the proposed methods assume that the true propensity score is known. 

\section{Cohort}
In this section, we describe how we define the cohort and process the data to be used in our proposed framework. 

\subsection{Cohort Selection}
The Medical Information Mart for Intensive Care database (MIMIC-III) is a freely accessible database, which contains data including $53,423$ Intensive Care Unit (ICU) admissions of adult patients between 2001 and 2012\cite{johnson2016mimic}. In this paper, we consider a cohort of patients from MIMIC-III v1.4, who fulfill the Sepsis-3 criteria\cite{singer2016third}. We follow the scripts\footnote{\url{https://github.com/matthieukomorowski/AI\_Clinician}} provided by Komorowski et al.\cite{komorowski2018artificial} to recreate the cohort. In short, the inclusion criteria select those adult patients who are associated with a Sequential Organ Failure Assessment (SOFA) score of 2 or more during the time of interest. The SOFA score ranges from 0 to 24, and a higher value indicates a more severe status of the patient. Further, patients with extreme unusual records or death during the data collection period are excluded from the cohort, as their records would have led to spurious policies. In total, $20,944$ admissions are included in our dataset. 

\subsection{Data Description and Processing}\label{sec: dataprocess}
\textbf{Static and sequential information}: There are two classes of variables that are relevant for modeling the treatment decisions: 1) static information, e.g., age and gender; 2) sequential information, e.g., time-varying heart rate and respiratory rate. Similar to Komorowski et al.\cite{komorowski2018artificial}, we extract a set of 47 variables, including information about demographics, vital signs, and lab values. Three of those variables are about static information and 44 are about sequential information. More details about the variables can be found in Appendix~\ref{app: feature}. The time of interest is defined as 24 hours before the onset of the sepsis and 48 hours after it. To represent the sequential status of the patient, we aggregate the data by averaging over four-hour windows. As a result, at each time-step, each admission is represented by a multidimensional vector. 

\textbf{Treatments and outcome}: We choose to learn the optimal ITRs for the administration of IV and VP, considering the suboptimality of their administration reported in the clinical literature\cite{byrne2017fluid}. More specifically, we follow the scripts from Komorowski et al. and define 25 treatment decisions for each four-hour time window, where each decision is an IV-VP pair for discretized dosages. The original dosage is first converted to zero (i.e., zero dosage) and non-zero classes, and the non-zero classes are further divided into quartiles. More statistics of the discretized treatment decisions can be found in Appendix~\ref{app: treatment}.

In the bandit problem, each action immediately receives feedback information. Therefore, we compute a clinically guided outcome, denoted by $\Delta$-SOFA  (differences between subsequent SOFA scores), as our feedback information to guide the learning of the ITRs. As concluded by Vincent et al.\cite{ferreira2001serial}, the $\Delta$-SOFA offers an objective evaluation of treatment responses and could be used to reflect patients' responses to therapeutic strategies. Furthermore, if a patient has an unchanged SOFA score in a low range or a decreased SOFA score in subsequent time windows, he/she is associated with a lower mortality rate. Similar applications of the $\Delta$-SOFA have been reported by Raghu et al.\cite{raghu2017deep}. In BLBF, the problem is cast as a risk minimization problem. Thus, we define the loss as $0$ (positive outcome) if $\Delta$-SOFA is unchanged in a low SOFA range (0-5) or has decreased. Otherwise, we set the loss as $1$ (negative outcome). 

\begin{figure}[!htb]
    \centering  
    \includegraphics[width=0.48\textwidth]{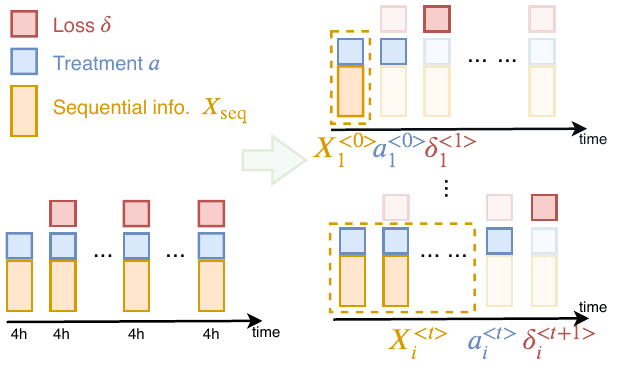}
    \caption{Illustration of the training and test sample generation from the medical history of each admission: The left-hand side presents the raw data after aggregating for every four-hour time window; the right-hand side shows the generated training and test samples. 
    A sample will be extracted if the following two conditions are fulfilled: 1) A treatment decision is observed at a certain time-step. 
    2) The feedback information is observed in the following time-step of the treatment. To highlight the relative order between the sequential information, treatment, and the corresponding loss, we add the superscript $< t >$ to indicate the time-step index of the treatment during the admission. \label{fig: trajectory}}
\end{figure}

\textbf{Training and test sample generation}: To model the treatment decision, we extract samples from the patients' medical history in an expanding window fashion, whenever a treatment is observed. For predictive modeling, the treatment decision is viewed as the target variable for training. All sequential information before the treatment is used as covariates for prediction. The sequential information at the time-step of the treatment is not used for learning, as some variables may not be observed at the time of decision in the ICU.
For ITRs learning, the outcome for the treatment decision is required. Therefore, we extract the observed $\Delta$-SOFA in the next time-step and compute the corresponding loss as described in the previous paragraph. As shown in Fig.~\ref{fig: trajectory}, each sample consists of the sequential information, treatment decision, and the corresponding loss information. In addition, the static information is also extracted but not shown in the figure for the sake of simplicity. 

From $20,944$ admissions we could extract in total $224,333$ samples (i.e., $10.7$ samples per admission on average). The number of time-steps observed before the treatment varies from $1$ to $18$ and is on average $7.2$. When generating the training samples and test samples, the split is based on the admission level rather than the sample level so that we can achieve a more objective evaluation. With the split admissions, the samples are divided for training and testing accordingly.

\section{Method}
Our proposed framework consists of two consecutive parts: a predictive model for the propensity score estimation and an ITRs model trained with an objective function based on the estimated propensity score. After following the preprocessing steps in Fig.~\ref{fig: trajectory}, we denote our data as $\{(X_\text{seq})_i, (\bm{x}_\text{sta})_i, a_i, \delta_i\}_{i=1}^m$, where $X_\text{seq}\in\mathbb{R}^{T\times 44}$ represents the (multivariate) random variable for the sequential information with $T$ observed time-steps and $\bm{x}_\text{sta}\in\mathbb{R}^3$ stands for the static information. We denote the treatment decision as $a\in \{1, 2, \dots, 25\}=: \mathcal{A}$ and the loss of the observed treatment as $\delta\in\{0, 1\}$. Scalars are denoted by lowercase letters such as $a, \delta$; (column) vectors are denoted by bold lowercase letters such as $\bm{x}_\text{sta}$; matrices are denoted by uppercase letters such as $X_\text{seq}$; sets are denoted by calligraphic letters such as $\mathcal{A}$.

\subsection{Propensity Score-Based Objective Function for Learning ITRs }\label{sec: psb}

Following the formulation in BLBF, the goal of learning the optimal ITRs is to find a policy $\pi_{\bm{w}}$ that minimizes the risk 
\begin{align} 
r(\pi_{\bm{w}}) 
&= \mathbb{E}_{X\sim \mathbb{P}(X)}\mathbb{E}_{a\sim \pi_{\bm{w}}(a|X)}\big[ \delta(X, a) \big]  \nonumber  \\
&=  \mathbb{E}_{X\sim \mathbb{P}(X)}\mathbb{E}_{a\sim \mathbb{P}(a|X)}\Big[ \delta(X, a)\cdot  \frac{\pi_{\bm{w}}(a|X)}{\mathbb{P}(a|X)} \Big] \label{eq: risk}
\end{align}
where $\bm{w}$ denotes the parameters of the policy. The loss $\delta(X, a)$ is an indicator function, which is 1 for negative outcome and 0 for positive outcome. 
The propensity score is reflected in the conditional probability $\mathbb{P}(a|X)$ for different treatments $a\in\mathcal{A}$. For conciseness, we use $X$ to denote the random variable for the complete medical history, including the sequential information $X_\text{seq}$ and the static information $\bm{x}_\text{sta}$, though it is a slight abuse of notation.. 

Equation~\eqref{eq: risk} is derived by applying importance sampling to remove the distribution mismatch between the physicians' policy and the new policy $\pi_{\bm{w}}$ . Intuitively, the new policy $\pi_{\bm{w}}$ will have a lower expected risk $r(\pi_{\bm{w}})$ when 
it has a higher probability for treatments with positive outcomes and a lower probability for treatments with negative outcomes.

The Inverse Propensity Score (IPS) estimator
\begin{equation}\label{eq: ips}
\hat{r}_\text{IPS}(\pi_{\bm{w}}) = \frac{1}{m} \sum^m_{i=1} \delta_i \frac{\pi_{\bm{w}}(a_i|X_i)}{\mathbb{P}(a_i|X_i)}
\end{equation}
applies Monte Carlo sampling to estimate the expected risk in \eqref{eq: risk} by taking the observed data points as samples. 
The IPS estimator will be unbiased if $\mathbb{P}(a_i|X_i)$ describes the physicians' policy. Therefore, it is appealing to use the risk defined by the IPS estimator (IPS risk) as the objective function to learn the optimal ITRs. 

However, there are mainly two reasons why it is not possible to optimize the policy using the IPS risk directly. First, it has been shown that the IPS estimator suffers from large variance if there is a large discrepancy between the new policy and the physicians' policy\cite{dudik2011doubly}, which would be more severe for high-capacity models like neural networks, as it is in our case. Second, directly minimizing an IPS estimator that contains the propensity score is prone to propensity score overfitting \cite{swaminathan2015self}. More specifically, the new policy is dominated by the physicians' policy rather than the treatment with low loss. 
In our setting, the minimal IPS risk in \eqref{eq: ips} is $0$.
The new policy will simply put zero probability on all the treatment decisions observed from the physicians. In other words, the new policy achieves minimal IPS risk by recommending any treatment that differs from the physicians' decision. In Sec.~\ref{sec: exp}, this phenomenon will also be empirically verified. 

Propensity score overfitting originates from \textit{the lack of equivariance} of the IPS estimator (see Appendix~\ref{app: equips}), i.e., the minimizer of the IPS risk is dependent on the translation of the loss. Furthermore, the lack of equivariance is due to the unconstrained treatment matching factor (TMF), defined as 
\begin{equation}\label{eq: TMF}
    s(\pi_{\bm{w}}) \coloneqq \frac{1}{m} \sum^m_{i=1} \frac{\pi_{\bm{w}}(a_i|X_i)}{\mathbb{P}(a_i|X_i)}
\end{equation}
which equals to $1$ in expectation (see Appendix~\ref{app: TMF}), but will be far from $1$ if the propensity score overfitting problem occurs.

As a solution, the self-normalized IPS estimator (SNIPS) 
\begin{equation}\label{eq: snips}
\hat{r}_\text{SNIPS}(\pi_{\bm{w}}) = \frac{\frac{1}{m} \sum^m_{i=1} \delta_i \frac{\pi_{\bm{w}}(a_i|X_i)}{\mathbb{P}(a_i|X_i)}}{s(\pi_{\bm{w}})} 
\end{equation}
is proposed to replace the IPS estimator for the learning of a new policy \cite{swaminathan2015self}. 
It is proven to be asymptotically unbiased \cite{hesterberg1995weighted} and has the property of equivariance (see Appendix~\ref{app: equsnips}), which enables the new policy to focus on learning the treatment with low loss.

Neural networks are typically trained by mini-batch stochastic gradient descent. Unfortunately, the optimization problem, including the SNIPS estimator, cannot be solved directly by a mini-batch stochastic gradient descent-based method, since all samples are required to compute the denominator. 
A mini-batch of samples could be used to estimate it, but the result is proven to be biased\cite{joachims2018deep}. Joachim et al. propose the \textit{BanditNet} by reformulating the SNIPS estimator with an additional constraint\cite{joachims2018deep}. 
In short, optimizing the SNIPS estimator is equivalent to optimizing a $\lambda$-translated IPS estimator 
\begin{equation} \label{eq: tips}
    \hat{r}^{\lambda}_\text{IPS}(\pi_{\bm{w}}) =
    \frac{1}{m} \sum^m_{i=1} (\delta_i-\lambda) \frac{\pi_{\bm{w}_j}(a_i|X_i)}{\mathbb{P}(a_i|X_i)}
\end{equation}
where the Lagrange multiplier $\lambda$ is called the \textit{translation} (more details in Appendix~\ref{app: reformulation}). The optimal translation $\lambda$ is found through grid search. As mentioned earlier, a translation of the loss results in a difference among the minimizers of the IPS risk: On the one hand, the new policy tends to avoid the treatments in the physicians' policy if losses are defined as non-negative values; on the other hand, it prefers to over-present the physicians' policy if losses are defined as non-positive values. Taking advantage of the lack of equivariance of the IPS estimator, the reformulation searches the optimal translation to balance these two tendencies so that the policy can focus on learning the treatment with low loss. 
\subsection{Predictive Modeling of Treatment Decisions}\label{sec: pm}
In observational studies, the propensity score is not known but can be estimated from the collected data. More specifically, the propensity score can be modeled by any supervised machine learning models that provide probability estimates for the various treatment decisions. We propose to apply state-of-the-art predictive models to produce an estimated propensity score $\hat{\mathbb{P}} (a=a_i|X)$, which is necessary for the optimization problem in \eqref{eq: tips}.

Recurrent neural networks (RNNs) provide an extension of feedforward neural networks to handle sequential inputs. Formally, given an input sequence $(\bm{x}_1, \bm{x}_2, \dots,\bm{x}_T)$, an RNN calculates the hidden states $\bm{h}_t$ at time-step $t$ iteratively by joining the current input at $t$ and the previous hidden state at $t-1$ as
\begin{align} \label{eq:simple_RNN}
\bm{h}_t = g(W \bm{x}_t + U \bm{h}_{t-1})
\end{align}
where $g(\cdot)$ is a non-linear activation function and $W$ and  $U$ are parametric weight matrices.
Since each hidden state is again dependent on its predecessor, the state at $t$ is theoretically capable of storing all relevant information of the entire history.
Downstream models for classification or regression tasks could be implemented to consume the hidden state $\bm{h}_t$ as their input. However, the classical RNN architecture as in \eqref{eq:simple_RNN} often suffers from the vanishing gradient problem \cite{bengio1993problem, hochreiter1997long} and therefore could fail to capture the long-term dependencies from the previous inputs. More advanced variants of RNNs, such as gated recurrent unit (GRU)\cite{chung2014empirical} or long short-term memory (LSTM)\cite{hochreiter1997long}, have been proposed to solve the problem with gating mechanisms and have achieved great successes in modeling sequential data with long-term dependencies, such as texts or sensory data\cite{graves2013speech}. 

\begin{figure}[!htb]
    \centering  
    \includegraphics[width=0.45\textwidth]{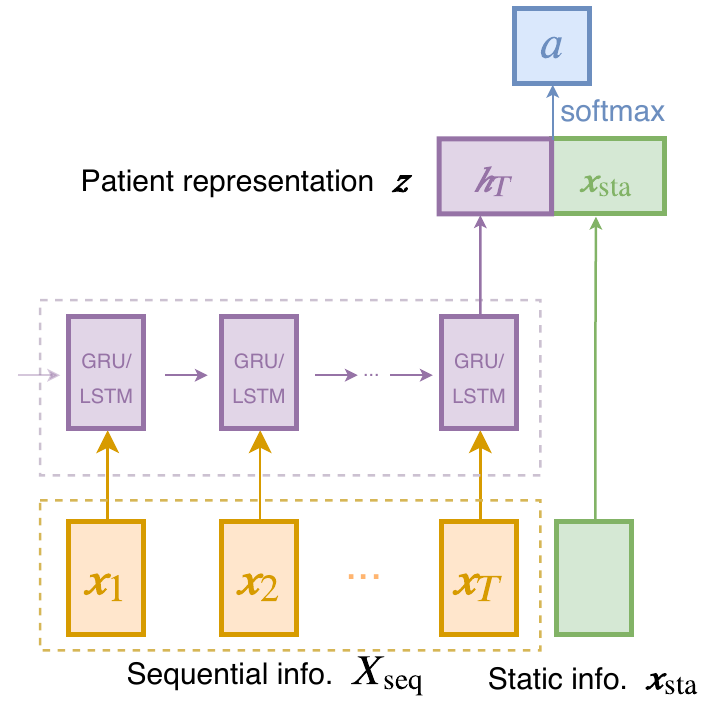}
    \caption{Illustration of the predictive modeling of treatment decisions with static and sequential information: GRU/LSTM encodes the sequential information into hidden states. The last hidden state is concatenated with the static information, resulting in a vector to represent the patient's complete medical history. On top of it, a softmax classifier is built to predict the treatment decisions of physicians. \label{fig: rnn}  }
\end{figure}

In the case of predictive modeling of treatment decisions, the multidimensional vector $\bm{x}_t$ at different time-steps constitutes the sequential input data $X_\text{seq}$. GRU/LSTM is used to encode $X_\text{seq}$ into the hidden states $\bm{h}_t$. Since we are mainly interested in modeling treatment decisions, a many-to-one structure is used\cite{yang2017predictive}, i.e., only the representation of the last hidden state $\bm{h}_T$ is utilized as the input for the treatment prediction, where $T$ is the number of observed time-steps before the treatment. Formally, we have 
\begin{align*}
    \text{GRU/LSTM}: \mathbb{R}^{T \times 44} & \to \mathbb{R}^h \\
    X_\text{seq} & \mapsto \bm{h}_T
\end{align*}
where $h$ is the dimension of the hidden state and will be tuned as a hyperparameter in the experiments. The static information is concatenated with the hidden state encoded by GRU/LSTM so that the static information is included for the modeling of the treatment decisions\cite{esteban2016predicting}. Formally, we have 
\begin{equation*}
    \bm{z} = (\bm{h}_T, \bm{x}_\text{sta}).
\end{equation*}
The resulting vector $\bm{z}\in \mathbb{R}^{h+3}$ represents the patient's complete medical history in a latent vector space and facilitates different subsequent tasks. In our case, a softmax classifier is built on top of it for the treatment prediction, as illustrated in Fig.~\ref{fig: rnn}.
In our framework, we interpret the probability distribution produced by this model as an estimate of the true propensity score. 

\subsection{Estimated Translated Inverse Propensity Score}
In this section, we elaborate the entire etIPS framework in Algorithm~\ref{alg: etips} by inversely joining the modules that have been introduced in Sec. \ref{sec: psb} and \ref{sec: pm}. 

In line~\ref{line: pm}, we train the predictive model as in Sec.~\ref{sec: pm} to estimate the physicians' policy. 
In line~\ref{line: computepi}, we derive the estimated propensity scores on all patient cases from the predictive model in line~\ref{line: pm}.
From line 3 to 6, we train our ITRs model as follows: 
For the $j$-th iteration, we select a particular translation $\lambda_j\in(0, 1)$ with grid search. The translation range is defined as $(0, 1)$ because the translation of $0$ makes all losses non-negative and the translation of $1$ makes all losses non-positive in our setting, which are the two extreme cases for the propensity score overfitting problem. We randomly initialize the trainable parameters $\bm{w}_j$ in the ITRs model, which has the same network structure as the predictive model in Fig.~\ref{fig: rnn} but is optimized with an objective function based on the estimated propensity score. 
Depending on the translation $\lambda_j$, we minimize the objective functions with respect to the trainable parameters $\bm{w}_j$ (line~\ref{line: baseline}). For each $\lambda_j$, both the minimizer $\bm{w}_j^*$ and its corresponding treatment matching factor $s_j$ (line~\ref{line: sj}) are saved. 
In line 7, the final minimization step outputs the pair $(s^*, \bm{w}^*)$ that generates the minimum value for the SNIPS risk in \eqref{eq: snips}. 

The differences between the minimization goal in line~\ref{line: baseline} and the IPS risk in \eqref{eq: ips} are the estimated propensity score $\hat{\mathbb{P}} (a=a_i|X)$ and the translation $\lambda_j$. Therefore, we name the algorithm \underline{e}stimated \underline{t}ranslated \underline{I}nverse \underline{P}ropensity \underline{S}core (etIPS). Intuitively, the proposed framework enables the new policy to be trained through encouraging the network to learn from the physicians' treatment decisions with a positive outcome as well as from unsuccessful cases (treatments with a negative outcome). 
\begin{algorithm}
	\SetKwInput{KwIn}{Input}
	\SetKwInput{KwOut}{Output}
    \SetAlgoLined
\KwIn{A dataset of the form $\{X_i, a_i, \delta_i\}_{i=1}^m$.}
\KwOut{The policy of the optimal ITRs $\pi_{\bm{w}^*}(a|X)$.
}
Learn the physicians' policy $\hat{\mathbb{P}} (a|X)$ with $\{X_i, a_i\}_{i=1}^m$ using the network structure in Fig.~\ref{fig: rnn}. \label{line: pm}\\
Compute the estimated propensity score $\hat{p}_i\coloneqq\hat{\mathbb{P}} (a=a_i|X_i)$ for all $i$. \label{line: computepi}\\
\For{$\lambda_j \in (0, 1)$}{
$\bm{w}^*_j \leftarrow
    \argmin_{\bm{w}_j}
    \bigg\{
    \frac{1}{m} \sum^m_{i=1} (\delta_i-\lambda_j) \frac{\pi_{\bm{w}_j}(a_i|X_i)}{\hat{p}_i}
    \bigg\}$  \label{line: baseline}\\
$s_j \leftarrow \frac{1}{m} \sum^m_{i=1} \frac{\pi_{\bm{w}_j^*}(a_i|X_i)}{\hat{p}_i}$ \label{line: sj}\\
}
$s^*, \bm{w}^* \leftarrow \argmin_{s_j, \bm{w}_j^*} \bigg\{ \frac{1}{s_j}\frac{1}{m} \sum^m_{i=1} \delta_i\frac{\pi_{\bm{w}_j}(a_i|X_i)}{\hat{p}_i} \bigg\} $\label{line: outermin}\\
\KwRet{$\pi_{\bm{w}^*}(a|X)$}
\caption{\textbf{etIPS}} 
\label{alg: etips}
\end{algorithm}

\section{Experiments}\label{sec: exp}
In this section, we provide details of the experiments conducted on three tasks, of which two are tailored to the BLBF setting from the MNIST dataset, which is common for the evaluation of many learning algorithms. As the ground truth labels in the MNIST dataset are available, the performance is evaluated with the metric accuracy. It serves as a simulation study\cite{zhou2017residual, swaminathan2015counterfactual, joachims2018deep}. In contrast, the MIMIC-III dataset only contains the feedback information of assigned treatments, and the offline evaluation is therefore employed, which can estimate the risk of a new policy from data observed from physicians.

\subsection{Implementation details}
The neural network-related models are built with the \textit{tensorflow} package\cite{tensorflow2015-whitepaper}. Hyperparameters are tuned with the \textit{hyperopt} package\cite{Bergstra_2015}. Five-fold cross-validation is implemented to report the variance of the performance.\footnote{Related scripts see \url{https://github.com/ZhiliangWu/etips}. }

\subsection{Simulation studies}
In this section, we simulate two controllable modeling tasks that resemble the true data situation. We aim to verify the following hypotheses empirically: a) Without ground truth labels, the propensity score-based objective function is applicable to sequential classification tasks. b) The estimated propensity score could be used to replace the true propensity score in the propensity score-based objective function. 
\subsubsection{Dataset generation}
We define two sequential classification tasks based on the MNIST dataset. The first task, \textit{zeros counting MNIST}, is to predict the number of zeros given a sequence of randomly sampled digit images. During sampling, we limit the maximum number of 0's to be 2 so that the prediction takes the form of a classification task with 3 classes. 
The second task, \textit{row-by-row MNIST}, is to predict the label of the digit ($0-9$) of the image. Each row of the image is presented sequentially to the neural network, and the classification is performed after reading all rows. Like other supervised learning tasks, the resulting dataset is in the form of $\{X_i, a^*_i\}_{i=1}^m$, where $a_i^*$ is the ground truth label. From that, \textit{the supervised to bandit conversion method}\cite{agarwal2014taming} is employed to generate BLBF datasets of the form $\{X_i, a_i, \delta_i, p_i\}_{i=1}^m$. If we view the tasks in a BLBF perspective, the context is a sequence of images $X$, the action $a$ is the label prediction of the given sequence, the loss $\delta$ reflects the correctness of the prediction, and $p$ is the probability of the label prediction. A \textit{logging policy}, which is similar to the physicians' policy in the clinical setting, is required to generate the label prediction for different contexts. Also, we need to set a suboptimal accuracy for it, just like we assume there is still improvement space for physicians. Similar to the conversion procedure\cite{swaminathan2015counterfactual}, we train a neural network to output $\mathbb{P}(a|\cdot)$ based on $5\%$ of the supervised dataset $\{X_i, a^*_i\}_{i=1}^m$ and select the one with an accuracy around $66\%$ as the logging policy. The label prediction $a_i$ is sampled from the output distribution of the logging policy. Meanwhile, the propensity score $p_i$ is also recorded for the sampled action. Finally, the loss $\delta_i$ is computed based on the ground truth label $a^*_i$, i.e., the loss is $0$ if the label prediction is the ground truth label and $1$ otherwise. More details of these generated datasets can be found in Appendix~\ref{app: mnist}. 
\subsubsection{Baselines}
For all approaches except the direct method, a many-to-one structure with GRU/LSTM is used to deal with the sequential inputs. The neural network structure in Fig.~\ref{fig: rnn} is not used because there is only sequential image information for the defined tasks. For the direct method, loss prediction is defined as the task for the network, and the action of label prediction is integrated in a way similar to the static information as in Fig.~\ref{fig: rnn}. 
\begin{enumerate}[a.]
    \item Direct method (DM): This method splits the task into two steps: It first learns the mapping $\mathbb{E}[\delta|X, a]$ to the expected loss given the context and action. The label prediction is then made by selecting the action with the lowest predicted loss $\argmin_a \mathbb{E}[\delta|X, a]$. 
    \item Random policy (RP): A dummy policy to perform a label prediction uniformly at random, which serves as a weak baseline. 
    \item Inverse Propensity Score (IPS): The network is trained to minimize the IPS risk as defined in \eqref{eq: ips}. 
    \item Translated Inverse Propensity Score (tIPS): The network is trained by minimizing the $\lambda$-translated IPS risk as defined in \eqref{eq: tips}. 
    \item Estimated Inverse Propensity Score (eIPS): The network is trained by minimizing the IPS risk as defined in \eqref{eq: ips} with the estimated propensity score.
\end{enumerate}

\subsubsection{Results} \label{exp: mnist}
Tab.~\ref{tab: mnist} shows the prediction performance of different approaches. Both tIPS and etIPS achieve more than $90\%$ accuracy, where etIPS yields the best results. RP has an accuracy of around $\frac{1}{\text{\#actions}}$, which is better than DM and IPS/eIPS.

\begin{table}[!htb]
\caption{Accuracy of different approaches on sequential classification tasks}
\label{tab: mnist}
\begin{minipage}{0.5\textwidth}
\centering
\begin{tabular}{@{}cccc@{}}
\toprule
 & propensity score & zeros counting MNIST & row-by-row MNIST \\ \midrule
DM & -\footnote{The propensity score is not involved in the algorithm.} & $0.343 \pm 0.0001$ & $0.098 \pm 0.0022$ \\
RP & -$^*$ & $0.363 \pm 0.0001$ & $0.103 \pm 0.0001$ \\ \cmidrule(l){2-4} 
IPS & true & $0.301 \pm 0.012$ & $0.020 \pm 0.0061$ \\
tIPS & true & $0.899 \pm 0.0229$ & $0.931 \pm 0.0852$ \\ \cmidrule(l){2-4} 
eIPS & estimated & $0.319 \pm 0.0075$ & $0.016 \pm 0.0098$ \\
etIPS  & estimated & $\bm{0.923 \pm 0.0122}$ & $\bm{0.953 \pm 0.0390}$ \\ \bottomrule
\end{tabular}
\end{minipage}
\end{table}

\subsubsection{Discussion}
Trained on partial feedback information, an optimal policy should also be able to perform the label prediction with the lowest risk (i.e., the highest accuracy), which works in the same way as an optimal classifier. Furthermore, accuracy serves as a good metric here to evaluate a new policy because the datasets have a balanced distribution for different output classes. The accuracy trained with cross-entropy loss and full label information is around $95\%$ for both tasks. From the performance of tIPS/etIPS, we see that the (estimated) propensity score-based objective function can deliver satisfying performance on the sequential prediction tasks when the ground truth label is not available. In addition, the performance of etIPS is a little better than tIPS. As proven\cite{xie2018off}, the reason is that the estimated propensity score has the potential to reduce the variance during the learning procedure. Furthermore, the performance of IPS/eIPS is worse than the weak baseline RP. Its poor performance is due to the propensity score overfitting problem, which can be diagnosed by computing the treatment matching factor in \eqref{eq: TMF}. For example, in the row-by-row MNIST task, $s(\pi_\text{IPS})=0.0061$ while $s(\pi_\text{tIPS})=0.926$. Last but not least, the performance of DM is as poor as RP. In practice, the performance for modeling $\mathbb{E}[\delta|X, a]$ is good with an accuracy of more than $85\%$ ($0.843 \pm 0.0072$ and $0.888 \pm 0.0027$ respectively). However, for the loss prediction, the network is trained with only one action under a certain context. The knowledge of the losses of different actions under the same context is missing during  training. As a result, the trained network would predict similar loss values for different actions under the same context, which accounts for the poor performance on the prediction task.

\subsection{Experiments on the MIMIC-III dataset}
\begin{table*}[!htb]
\caption{Evaluation with different risk estimators }
\label{tab: evaluation}
\centering
\begin{minipage}{\textwidth}
\centering
\begin{tabular}{@{}cccc|c@{}}
\toprule
 & ATENP & IPS& DR & TMF\\ \midrule
Predictive Modeling & $-0.019 \pm 0.0021$ & $0.523 \pm 0.0229$ & $0.523 \pm 0.0021$ & $1.034 \pm 0.0391$ \\
Direct Method\footnote{There is no probability of the treatment suggestion given by $\argmin_a\mathbb{E}[\delta|X, a]$. The values for IPS/DR/TMF can therefore not be computed.} & $0.032 \pm 0.0001$ & - & - & - \\
Most Frequent\footnote{A deterministic policy to suggest the most frequent treatment. There is no probability information involved.} & $-0.023 \pm 0.0001$ & - & - & - \\
Random Policy & $-0.023 \pm 0.0001$ & $0.125 \pm 0.0001$ & $0.478 \pm 0.0026$ & $0.243 \pm 0.0001$ \\
Estimated Inverse Propensity Score & $-0.025 \pm 0.1009$ & $\bm{0.009 \pm 0.0019}$ & $0.504 \pm 0.0071$ & $0.018 \pm 0.0029$ \\
Estimated Translated Inverse Propensity Score & $\bm{-0.143 \pm 0.0099}$ & $0.169 \pm 0.0160$ & $\bm{0.471 \pm 0.0060}$ & $0.438 \pm 0.0279$ \\ \bottomrule
\end{tabular}
\end{minipage}
\end{table*}
\subsubsection{Evaluation metrics}\label{subsec: evaluation}
For the MIMIC-III dataset, the goal is to learn the optimal ITRs for the administration of IV and VP. It is worth mentioning that offline evaluation remains a challenge, and the new policy requires further investigation with domain experts like physicians\cite{gottesman2018evaluating}. A new policy is hereby evaluated with three different evaluation methods:
\begin{enumerate}[a.]
    \item Average Treatment Effects under the new policy (ATENP): This method evaluates the new policy in a deterministic way, i.e., it only considers the treatment suggestion with the highest probability. According to the treatment suggestions of the new policy, the samples in the test set are divided into two groups: those who follow the new policy (group one) and those who do not (group two)\cite{zhou2017residual, komorowski2018artificial}. The difference between the average risk in these two groups shows the average treatment effects under the new policy. If a new policy is better than the physicians' policy, the difference should be below zero.\label{abb: atenp}
    \item Inverse Propensity Score Estimator (IPS): This method estimates the risk of the new policy as in \eqref{eq: ips}. As we discussed earlier, it may suffer from propensity score overfitting problem and thus be strongly biased. \label{abb: ips}
    \item Doubly Robust Estimator (DR): The \textit{doubly robust} technique consists of an outcome prediction model and a propensity score model\cite{dudik2011doubly} as \label{abb: dr}
    \begin{align*}
        \hat{r}_\text{DR}(\pi_{\bm{w}}) = \frac{1}{m}\sum_{i=1}^m \bigg[ &\sum_{a\in\mathcal{A}}\pi_{\bm{w}}(a|X_i)\hat{\delta}(X_i, a) \\
        &+ \frac{\pi_{\bm{w}}(a_i|X_i)}{\hat{\mathbb{P}}(a_i|X_i)}\big(\delta_i - \hat{\delta}(X_i, a_i)\big) \bigg]
    \end{align*}
    where $\hat{\delta}(X, a)$ is the loss prediction model and $\hat{\mathbb{P}}(a|X)$ is the propensity score model. It protects the mismodeling of either model by combining them to get the best of both.
\end{enumerate}
As the problem is formulated as a risk minimization problem, a lower value of ATENP/IPS/DR is preferred. In Appendix~\ref{app: evaluation}, the investigation of the correlation between accuracy and the risk estimated by these methods is provided to shed some light on the performance of different evaluation approaches. In short, ATENP shows a consistent correlation with the accuracy and is therefore trustworthy when there is a large sample size in group one. In comparison, the IPS estimator will be strongly biased when the propensity score overfitting problem occurs. In such cases, the DR estimator is more reliable by taking advantage of an outcome prediction model for correction. 

\subsubsection{Baselines}
The true propensity score is not available in observational studies, which prevents the application of IPS and tIPS. Instead, we implement DM, RP, and eIPS for evaluation purposes. The network structure for eIPS follows the one in Fig.~\ref{fig: rnn}, and DM is defined as a loss prediction task with the treatment as an additional input feature. In addition, the predictive modeling of treatment decisions (cf. Sec.~\ref{sec: pm}) and the most frequent policy are also included as baselines. The most frequent policy always suggests the most frequent treatment in the training dataset. In our case, it is the zero dosage of both IV and VP. Besides the evaluation methods, the treatment matching factor (TMF, cf. \eqref{eq: TMF}) is computed based on the estimated propensity score to diagnose the propensity score overfitting problem. 

\subsubsection{Results}
Tab.~\ref{tab: evaluation} shows the performance of the policies trained by different approaches. Our proposed approach turns out to have the lowest value in ATENP and DR with a high value of TMF (only lower than predictive modeling). In addition, the eIPS have the lowest value in the IPS evaluation with the lowest TMF.

\subsubsection{Discussion}
From a methodological perspective, the baseline approaches DM and eIPS can be viewed as the deep learning variants of the two-step method proposed by Qian et al. \cite{qian2011performance} and outcome weighted learning (OWL)\cite{zhao2012estimating}, respectively. Similarly, our proposed method can be understood as a deep learning variant of residual weight learning (RWL) proposed by Zhou et al. \cite{zhou2017residual}. The difference is that instead of learning a baseline by a separate regression model, our method is more efficient by trying different translations $\lambda_j$ to find the optimal baseline. Furthermore, a predicted baseline in RWL inevitably introduces additional noise in the loss, which can potentially deteriorate the learning. 

For ATENP, a value below zero means that the new policy is better than the physicians' policy. In the predictive modeling setting, the policy tries to mimic the physicians' policy as well as possible. The ATENP of it being around zero is therefore expected as it doesn't consider the outcome information. In comparison, the ATENP of etIPS shows a strong negative value of $-0.143$. It indicates that the observed treatments, which are the same as suggested by the new policy, have a much lower risk than those that are not. Also, the risk in group one of etIPS is estimated by $1929.8 \pm 111.33$ samples, which is relatively large, compared to $21.8 \pm 9.62$ for eIPS and $347 \pm 0.01$ for DM. 

The lowest IPS risk for the eIPS is strongly biased, which can be indicated by both the small sample size in group one ($21.8 \pm 9.62$) for ATENP and its lowest treatment matching factor ($0.018 \pm 0.0029$). The DR estimator corrects the bias with an outcome prediction model, resulting in a change from $0.009$ to $0.504$.

Last but not least, although the TMF value of etIPS is larger than other baselines except the predictive model, it is still a bit away from the expected value of $1$ (cf. Appendix~\ref{app: TMF}). Two reasons account for it. The first is the suboptimal accuracy of the predictive model, which is $0.571 \pm 0.0037$ in the test set. As mentioned earlier, the estimated propensity score is used to compute TMF. Therefore, it indicates the alignment between the policies of the predictive model and other models. As the predictive model cannot perfectly reflect the physicians' policy, the TMF value computed based on it does not necessarily have to be strictly around $1$ anymore. Nevertheless, the TMF computed from the predictive model is still worth being referenced when the value is extremely low like for eIPS. The second reason is the average risk in the dataset being $0.498$. In other words, almost half of the time, the physicians' treatment does not receive a positive outcome. The relatively large amount of negative feedback encourages the algorithm to learn a new policy that is a bit different from the physicians. Besides, there are 25 treatment decisions observed with strong skewness in its distribution (cf. Appendix~\ref{app: treatment}). These facts would jointly result in a lower TMF.

\section{Conclusion}
In this paper, we propose a general framework, etIPS, to learn optimal ITRs. It consists of a predictive model and an ITRs model. The former takes advantage of the state-of-the-art predictive modeling of the treatment decisions while the latter is based on the latest formulation of BLBF problems. By casting the ITRs learning as a problem in BLBF, our proposed approach can discover the optimal policies with sequential EHRs from observational studies. Intuitively speaking, the new policy is learned by encouraging the treatments with a positive outcome and discouraging the treatments with a negative outcome. The reformulation of the SNIPS estimator ensures that such a learning objective is correctly integrated into the objective function of the neural network. The generality of our proposed framework lies in the flexibility to choose an arbitrary propensity score model as well as any ITRs model that would fit the patient features. 

With experiments on two simulated BLBF tasks using the MNIST dataset, we have empirically shown that the estimated propensity score can replace the true propensity score when the latter is not known. The result facilitates the usage of data from observational studies without any recorded propensity score. Furthermore, in various offline evaluation methods, our learned policies perform better than the physicians' policy.  A true performance evaluation, naturally, would require additional clinical testing. 

The proposed framework is compatible with any neural network structures and any data sources, not limiting to recurrent neural networks and sequential EHRs, as we have presented in this paper. With more advanced network structures, the performance of our framework could be further boosted. 

As part of  future work, we want to study model explainability or interpretability. If the treatment suggestion is provided together with explanations, the physicians would find such clinical decision support systems more transparent and become more encouraged to apply it. For example, the explanation can show which parts of the static or sequential information are especially important for the final treatment suggestion.

\section*{Acknowledgment}
The authors acknowledge support by the German Federal Ministry for Education and Research (BMBF), funding project “MLWin” (grant 01IS18050).
\begin{figure}[!htb]
    \includegraphics[width=0.6\linewidth]{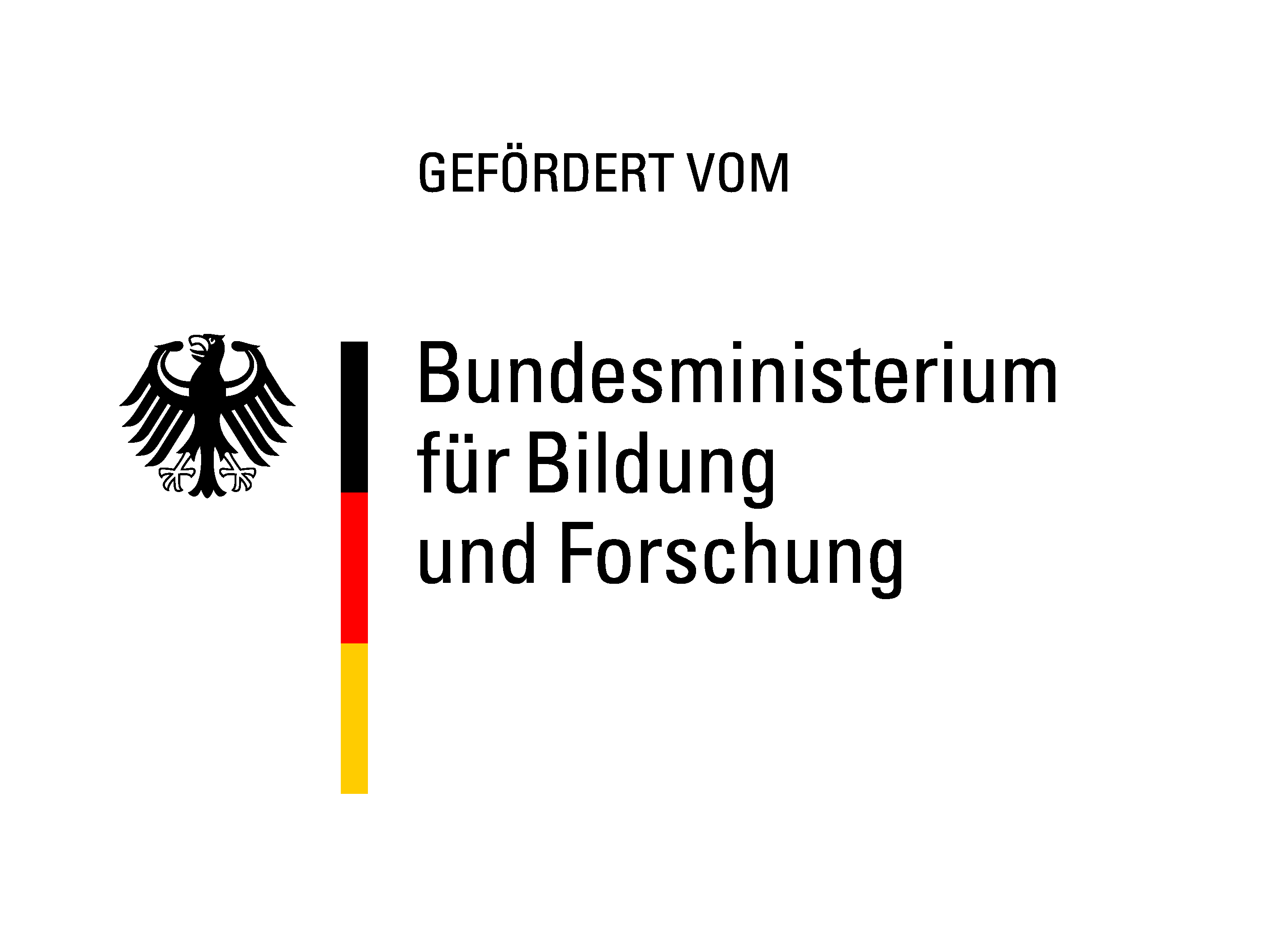}
\end{figure}

\bibliographystyle{IEEEtran}
\bibliography{Ref_ICHI2020}

\appendix
\subsection{Feature description}\label{app: feature}
The included features are chosen to best represent the status of each patient\cite{komorowski2018artificial}. There could possibly be confounding effects if we haven't included some important features in the model. In the chosen features, most have continuous values except for gender, readmission, and mechanical ventilation being binary. 

\textbf{Static information}: age, gender, readmission to intensive care.

\textbf{Sequential information}: weight (kg), Glasgow Coma Scale (GCS), heart rate(HR), Systolic, Mean and Diastolic Blood Pressure(SysBP, MeanBP, DiaBP), Respiratory Rate (RR), SpO2, temperature (celsius), FiO2, Potassium, Sodium, Chloride, Glucose, Blood Urea Nitrogen (BUN), Creatinie, Magnesium, Calcium, SGOT, SGPT, Total Bilirubin, Hemoglobin, count of the white blood cells, count of the platelets, Partial Thromboplastin Time (PTT), Prothorombin Time (PT), International Normalized Ratio (INR), Arterial potential Hydrogen, paO2, paCO2, Arterial Base Excess, Artial lactate, HCO3, mechanical ventilation, shock index, PaO2/FiO2 ratio, maximum dose of vasopressor over 4 hours, intravenous fluids intake over 4 hours, total input,  total urine fluid output, urine output over 4 hours, cumulated fluid balance, Sequential Organ Failure Assessment (SOFA) over 4 hours, Systemic Inflammatory Response Syndrome (SIRS) over 4 hours.
\subsection{Treatment decisions}\label{app: treatment}
\begin{figure}[!htb]
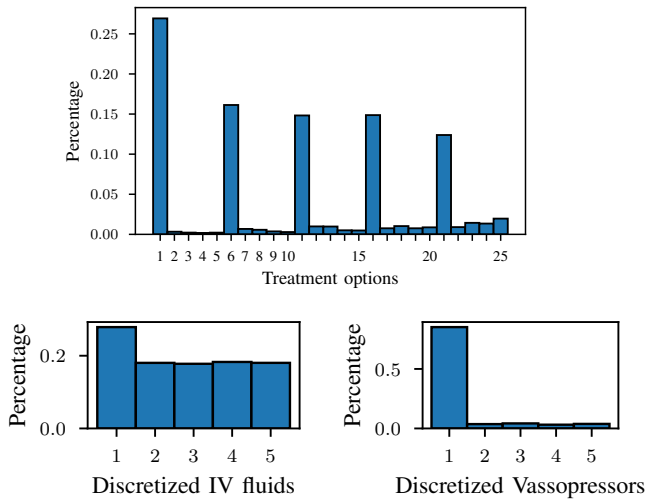

    \centering  
    \scalebox{0.7}{\input{treatment_options.pgf}}
      \begin{minipage}{0.2\textwidth}
    \centering
    \scalebox{0.9}{\input{IV_fluids_half.pgf}}
  \end{minipage}
  \qquad
  \begin{minipage}{0.2\textwidth}
    \centering
    \scalebox{0.9}{\input{Vassopressors_half.pgf}}
  \end{minipage}
    \caption{Distribution of treatment decisions and discretized IV/VP\label{fig: treat}}
\end{figure}

Fig.~\ref{fig: treat} shows the distribution of different treatment options. The skewness of the distribution is mainly due to the unbalanced distribution of the discretized VP. 


Due to the possible update of the MIMIC-III database, there are slight differences between the values in Table~\ref{tab: range} compared to the ones reported by Komorowski et al.\cite{komorowski2018artificial}.
\begin{table}[!htb]
\caption{Range and median of IV and VP}
\label{tab: range}
\begin{tabular}{|c|c|c|c|c}
\hline
 & \multicolumn{2}{c|}{IV fluids (mL/ 4 hours)} & \multicolumn{2}{c|}{Vasopressors (mcg/kg/min)} \\ \cline{2-5} 
Treatment & range & median & range & \multicolumn{1}{c|}{median} \\ \hline
1 & 0 & 0 & 0 & \multicolumn{1}{c|}{0} \\ \hline
2 & 0 - 48 & 30 & 0 - 0.08 & \multicolumn{1}{c|}{0.04} \\ \hline
3 & 48 - 150 & 80 & 0.08 - 0.2 & \multicolumn{1}{c|}{0.13} \\ \hline
4 & 150 - 500 & 284 & 0.2 - 0.45 & \multicolumn{1}{c|}{0.27} \\ \hline
5 & $>500$ & 874 & $ > 0.45$ & \multicolumn{1}{c|}{0.78} \\ \hline
\end{tabular}
\end{table}
\subsection{Treatment matching factor}\label{app: TMF}
\begin{align*}
&\mathbb{E}_{X\sim \mathbb{P}(X)}\mathbb{E}_{a\sim \mathbb{P}(a|X)}\bigg[\frac{\pi_{\bm{w}}(a|X)}{\mathbb{P}(a|X)}\bigg] \\
&=\sum_X \mathbb{P}(X)\sum_a \mathbb{P}(a|X)
 \frac{\pi_{\bm{w}}(a|X)}{\mathbb{P}(a|X)}\\
 &=\sum_X \sum_a \mathbb{P}(X)\pi_{\bm{w}}(a|X)\\
 &=1
\end{align*}
\subsection{Lack of equvariance of the IPS estimator}\label{app: equips}
$$
\min_{\bm{w}}\frac{1}{m} \sum^m_{i=1} (\delta_i + c) \frac{\pi_{\bm{w}}(a_i|X_i)}{\mathbb{P}(a_i|X_i)} \neq 
c + \min_{\bm{w}}\frac{1}{m} \sum^m_{i=1} \delta_i \frac{\pi_{\bm{w}}(a_i|X_i)}{\mathbb{P}(a_i|X_i)}
$$
\subsection{Equvariance of the SNIPS estimator}\label{app: equsnips}
\begin{align*}
&\min_{\bm{w}} \frac{\frac{1}{m} \sum^m_{i=1} (\delta_i + c) \frac{\pi_{\bm{w}}(a_i|X_i)}{\mathbb{P}(a_i|X_i)}}{\frac{1}{m} \sum^m_{i=1} \frac{\pi_{\bm{w}}(a_i|X_i)}{\mathbb{P}(a_i|X_i)}}  \\
&= \min_{\bm{w}} \bigg(\frac{\frac{1}{m} \sum^m_{i=1} \delta_i  \frac{\pi_{\bm{w}}(a_i|X_i)}{\mathbb{P}(a_i|X_i)}}{\frac{1}{m} \sum^m_{i=1} \frac{\pi_{\bm{w}}(a_i|X_i)}{\mathbb{P}(a_i|X_i)}}  + c \cdot  \frac{\cancel{\frac{1}{m} \sum^m_{i=1} \frac{\pi_{\bm{w}}(a_i|X_i)}{\mathbb{P}(a_i|X_i)}}}{\cancel{\frac{1}{m} \sum^m_{i=1} \frac{\pi_{\bm{w}}(a_i|X_i)}{\mathbb{P}(a_i|X_i)}}}\bigg) \\
&= \min_{\bm{w}} \frac{\frac{1}{m} \sum^m_{i=1} \delta_i  \frac{\pi_{\bm{w}}(a_i|X_i)}{\mathbb{P}(a_i|X_i)}}{\frac{1}{m} \sum^m_{i=1} \frac{\pi_{\bm{w}}(a_i|X_i)}{\mathbb{P}(a_i|X_i)}}
\end{align*}
\subsection{Reformulation of the SNIPS risk}\label{app: reformulation}
The optimization objective of the SNIPS risk
\begin{equation*}
    \bm{w}^* = \argmin_{\bm{w}}\frac{\frac{1}{m} \sum^m_{i=1} \delta_i \frac{\pi_{\bm{w}}(a_i|X_i)}{\mathbb{P}(a_i|X_i)}}{\frac{1}{m} \sum^m_{i=1} \frac{\pi_{\bm{w}}(a_i|X_i)}{\mathbb{P}(a_i|X_i)}}
\end{equation*}
could be reformulated as a two-step optimization problem 
\begin{align*}
    s^*, \bm{w}^* = \argmin_{s_j} \Bigg\{
    \argmin_{\bm{w}_j}
    \frac{\frac{1}{m} \sum^m_{i=1} \delta_i \frac{\pi_{\bm{w}_j}(a_i|X_i)}{\mathbb{P}(a_i|X_i)}}{s_j} 
    , \\
    \text{s.t. } \frac{1}{m} \sum^m_{i=1} \frac{\pi_{\bm{w}_j}(a_i|X_i)}{\mathbb{P}(a_i|X_i)} = s_j 
    \Bigg\}
\end{align*}
where $s_j$ is fixed to different values and $\bm{w}_j$ represents the corresponding optimization parameters. In other words, the minimizer can be found by 1) fixing $s_j$ to a particular value within a grid search, and 2) solving the corresponding interior constrained optimization problem to find $\bm{w}_j^*$. The final minimizer is the pair with the lowest SNIPS risk among all $(s_j, w^*_j)$ pairs.
 
The remaining problem is to solve the interior constrained optimization problem. It is natural to use the Lagrange multiplier to remove the constraint of the fixed $s_j$. Formally, the problem 
\begin{align*}
    \bm{w}^*_j =     \argmin_{\bm{w}_j}
    \bigg\{
    &\frac{1}{m} \sum^m_{i=1} \delta_i \frac{\pi_{\bm{w}_j}(a_i|X_i)}{\mathbb{P}(a_i|X_i)}
    , \\
    &\text{ s.t. } \frac{1}{m} \sum^m_{i=1} \frac{\pi_{\bm{w}_j}(a_i|X_i)}{\mathbb{P}(a_i|X_i)} = s_j 
    \bigg\}
\end{align*}
is equivalent to 
\begin{equation*}
    \bm{w}^*_j, \lambda_j^* =
    \argmin_{\bm{w}_j}\max_{\lambda_j}
    \bigg\{
    \frac{1}{m} \sum^m_{i=1} (\delta_i-\lambda_j) \frac{\pi_{\bm{w}_j}(a_i|X_i)}{\mathbb{P}(a_i|X_i)} + \lambda_j s_j
    \bigg\}.
\end{equation*}

Considering the fact that searching for $\lambda_j^*$ with a fixed $s_j$ is expensive but the inverse is not, reversing the role of $\lambda_j^*$ and $s_j$ makes the problem more tractable, i.e., fix $\lambda_j$ first, optimize for $\bm{w}_j^*$, and compute the corresponding $s_j$ as well as the SNIPS risk $\hat{r}_\text{SNIPS}(\pi_{\bm{w}_j^*})$. Formally, the optimization problem is further reduced to 
\begin{equation*}
    \bm{w}^*_j =
    \argmin_{\bm{w}_j}
    \bigg\{
    \frac{1}{m} \sum^m_{i=1} (\delta_i-\lambda_j) \frac{\pi_{\bm{w}_j}(a_i|X_i)}{\mathbb{P}(a_i|X_i)}
    \bigg\}.
\end{equation*}

\subsection{Sequential classification tasks from MNIST}\label{app: mnist}
\begin{figure}[!htb]
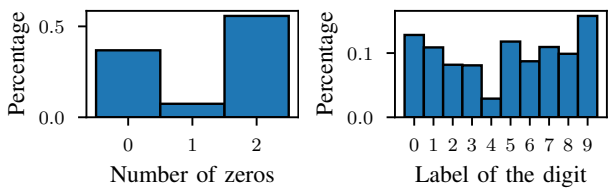

  \begin{minipage}{0.2\textwidth}
    \scalebox{0.9}{\input{counting_zeros.pgf}}
  \end{minipage}
  \quad
  \begin{minipage}{0.2\textwidth}
    \scalebox{0.9}{\input{row_by_row_mnist.pgf}}
  \end{minipage}
  \caption{Distribution of label prediction of the logging policy\label{fig: action_dist}}
\end{figure}

Table~\ref{tab: statm} shows some statistics for the tailored sequential classification tasks. The output classes in both tasks have a balanced distribution. Meanwhile, due to the preference of the logging policy, the label predictions show the skewness to some extend in Fig.~\ref{fig: action_dist}. 
\begin{table}[!htb]
\caption{Basic statistics of the tailored tasks with MNIST}
\label{tab: statm}
\centering
\begin{tabular}{|c|c|c|}
\hline
                                                                                & \text{zeros counting MNIST} & \text{row-by-row MNIST} \\ \hline
\#samples                                                                       & $10, 000$            & $70, 000$        \\ \hline
\begin{tabular}[c]{@{}c@{}}input shape\\ (\#time-steps, \#features)\end{tabular} & $(20\pm 5, 784)$     & $(28, 28)$       \\ \hline
\#output classes                                                                & $3$                  & $10$             \\ \hline
\end{tabular}
\end{table}

\subsection{Different evaluation methods}\label{app: evaluation}
\begin{figure}[!htb]
    \centering  
    \scalebox{0.7}{\input{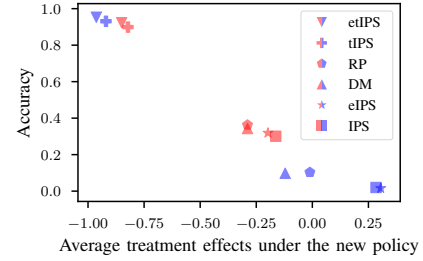}}
    \caption{Correlation between the accuracy and ATENP}
    \label{fig: acc-etenp}
\end{figure}

\begin{figure}[!htb]
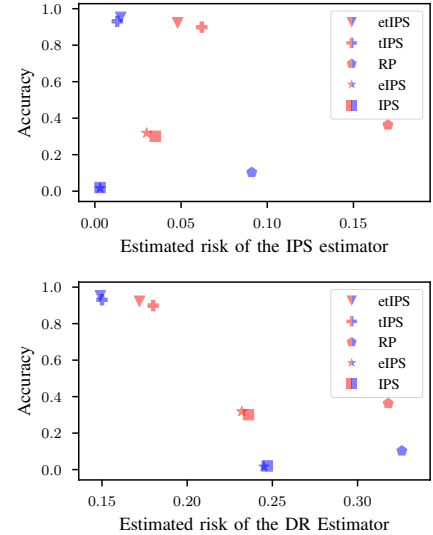

    \centering  
    \scalebox{0.7}{\input{IPS.pgf}}
    \scalebox{0.7}{\input{DR.pgf}}
    \caption{Comparison of the risk estimation with IPS/DR estimator}
    \label{fig: acc-ips-dr}
\end{figure}

As the accuracy is computed with the ground truth label, it serves as a good reference to understand the performance of different risk estimators. In Fig.~\ref{fig: acc-etenp} and \ref{fig: acc-ips-dr}, the blue color denotes the performance of zeros counting MNIST while red the row-by-row MNIST. 

Although the policy is evaluated in a deterministic way, ATENP shows a consistent correlation with the accuracy of different policies. In addition, the sample size in different groups serves as a good indicator for the propensity score overfitting problem. For the row-by-row MNIST, there are only $4.6\pm 3.83$ samples in group one for the policy learned with IPS, while the number is $4406.8 \pm 373.43$ for tIPS, which corresponds to the low treatment matching factors as discussed in Sec.~\ref{exp: mnist}.

In Fig.~\ref{fig: acc-ips-dr}, IPS/eIPS approaches have the smallest estimated risk, which indicates that the IPS estimator is strongly biased if the propensity score overfitting problem occurs. Taking advantage of an additional outcome prediction model, the DR estimator corrects the risk estimation and is therefore more reliable.

\end{document}